\documentclass[letterpaper, oneside, 10 pt, conference]{ieeeconf}
\IEEEoverridecommandlockouts

\usepackage[utf8]{inputenc}
\usepackage[T1]{fontenc}
\usepackage{amsmath}
\usepackage{amssymb}
\usepackage{capt-of}
\usepackage{hyperref}
\usepackage{amsmath}
\usepackage{amssymb}
\usepackage{pgfplots}
\usepackage{physics}
\usepackage{float}
\usepackage{tikz}
\usepackage{svg}
\usepackage{bm}
\usepackage{booktabs}
\usepackage{graphicx}
\let\labelindent\relax
\usepackage{enumitem}

\newcommand{\R}{\mathbb{R}}

\newcommand{\ninput}{x}

\newcommand{\blip}{\mathcal{G}}
\newcommand{\pln}{\mathcal{P}}
\newcommand{\nll}{z}

\newcommand{\cfg}{q}

\newcommand{\smm}{\mathcal{S}} 
\newcommand{\fwd}{f}
\newcommand{\task}{y}

\newcommand{\data}{\mathcal{D}}
\newcommand{\aux}{J}

\newcommand{\grav}{{\hat g}}

\newcommand{\methodnameacr}{{\text{LOInK}}}

\title{\LARGE \bf LOInK: Learned Optimal Inverse Kinematics\\via Structured Neural Surrogate Models
\thanks{The authors are with the Australian Robotic Inspection and Asset Management (ARIAM) Hub, the Australian Centre for Robotics (ACFR), and the School of Aerospace, Mechanical and Mechatronic Engineering, The University of Sydney, Australia {\tt \{michael.somerfield, damian.abood, ruigang.wang, ian.manchester\}@sydney.edu.au}.}}
\author{Michael Somerfield, Damian Abood, Ruigang Wang, Ian R. Manchester}

\begin{document}

\maketitle 

\global\csname @topnum\endcsname 0
\global\csname @botnum\endcsname 0

\begin{abstract}
We introduce  Learned Optimal Inverse Kinematics ($\methodnameacr$), a method to generate approximately optimal solutions to an inverse kinematics problem. When trained on data consisting of sampled configurations and associated task variables and a given cost function, $\methodnameacr$ learns a bi-Lipschitz invertible mapping from configuration space to a decoupled task/latent space, and moreover, the latent space is structured so as to place cost-minimizing solutions at the origin. This enables efficient sampling of cost-minimizing solutions via a network-inversion algorithm based on operator splitting.
We demonstrate the proposed approach on three problems: an illustrative three degree-of-freedom manipulator problem; a quadrupedal climbing robot for which LOInK can generate near-optimal solutions on average 31 times faster and up to 100 times faster than a constrained optimization approach; and a simulated soft actuator as a purely data-driven example, in which LOInK can explicitly generate high-quality solutions, unlike existing generative approaches that require diverse sampling and evaluation of candidate solutions. 

\end{abstract}

\mdseries

\section{Introduction}

\begin{figure}[t]
    \centering
    \includegraphics[width=0.9\linewidth]{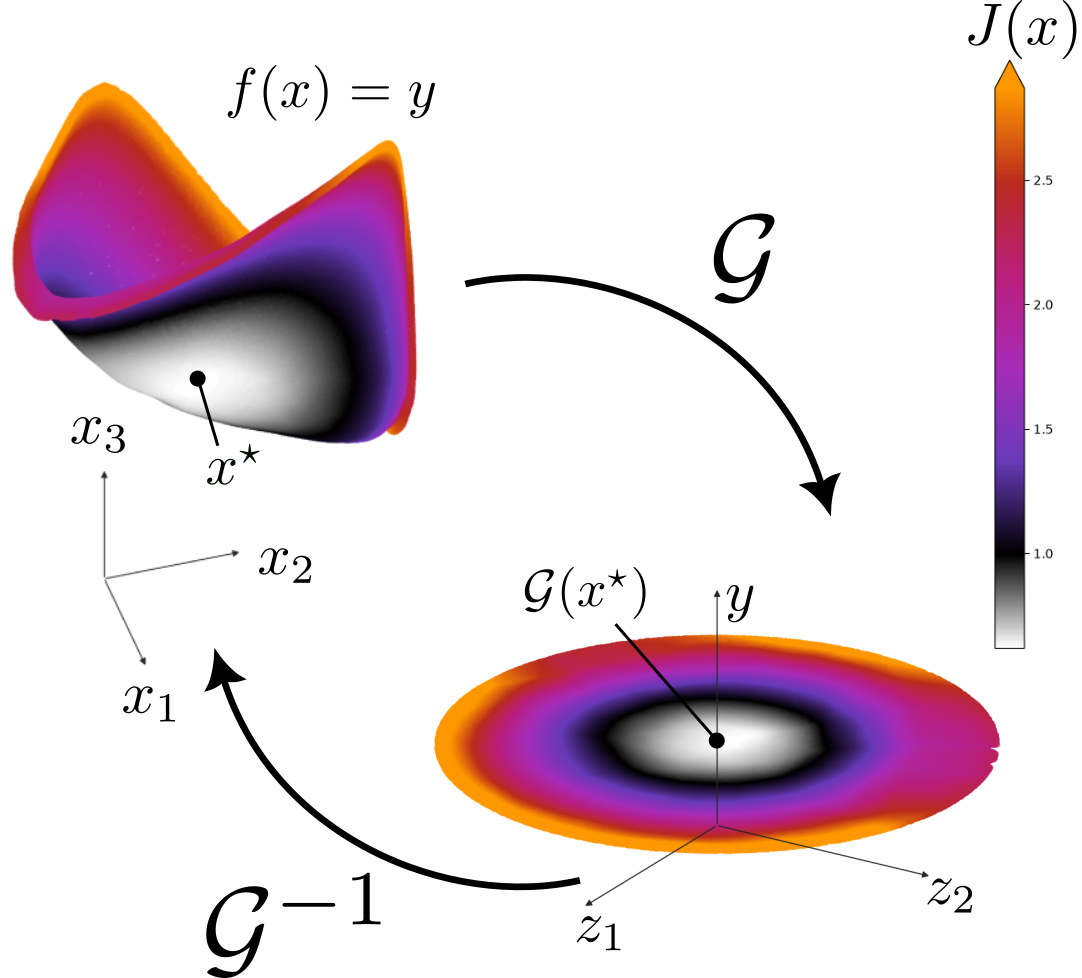}
    \caption{The main idea of LOInK is to learn an invertible map $\mathcal G$ (a bi-Lipschitz diffeomorphism) between configuration space and a decoupled task/latent-space coordinate system. Moreover, on the latent space, the origin $z=0$ corresponds to the cost-minimizing solution. Then, for a given task variable $y$, the optimal configuration can be evaluated as $x^\star=\mathcal G^{-1}([y;0])$.}
    \label{fig:title-figure}
\end{figure}

Inverse problems are significant in many fields, from medical imaging, geophysics, astronomy and others \cite{benning_modern_2018}, and as such are a focal point for many researchers. Finding solutions for these problems requires inferring which input parameters would produce a given output. For many inverse problems, there are multiple feasible solutions and it becomes necessary to choose between them, e.g. via regularization, prior knowledge or some measure of solution quality.

A prominent inverse problem within robotics is inverse kinematics (IK), where the question of how best to position a robot to achieve a desired task is ubiquitous across platforms, and for many situations remains challenging. For  a robot with redundancy, that is more degrees of freedom than task-space constraints, an IK method must choose one solution from a (generally infinite) set of solutions, variously referred to as a null space, a latent space, or a self-motion manifold~\cite{burdick1989inverse}. This provides freedom to select a solution which minimizes some auxiliary cost, e.g. manipulability metrics for manipulators, stability criteria for a legged platform, actuator effort, or costs related to collision avoidance. 

Analytical approaches to the problem are restricted to standard manipulator morphologies, and particular solutions are not necessarily high quality ~\cite{lakshmi_narayanan_decade_2026}. IK can be formulated as a constrained optimization problem, and standard numerical methods allow for locally searching the solution set for high quality solutions. However, IK problems can have challenging optimization landscapes, and this approach is fundamentally inapplicable when the forward model or cost function (or both) are computationally intractable for real-time optimization, or simply unavailable. 

Recently, learning-based methods have been introduced for the problem of positioning redundant manipulators, utilizing the null space afforded by redundancy to provide a set of solutions, as opposed to being restricted to a single solution as with non-redundant manipulators. Invertible networks have been utilized to learn the forward pass, as well as a decoupled null space. Once learned, the network's latent representation may be sampled, returning configurations which satisfy a desired task space position. These generative methods offer the ability to obtain one or many feasible solutions, however these approaches do not attempt to generate a solution that is high quality~\cite{xing_inverse_2026}.

To the best of the authors' knowledge, no learning-based method explicitly generates high-quality solutions to an inverse problem. The solutions are sampled arbitrarily from a distribution, and the cost must be evaluated on the solution/s after inference, which may be computationally expensive or simply impossible in a model-free setting. Further, there is no notion of smoothness between nearby task-space solutions, limiting applicability to time-varying task-space trajectories. 

In this paper, we introduce a new approach: Learned Optimal Inverse Kinematics ($\methodnameacr$), 
which learns an invertible change of coordinates to a structured task/latent space, such that for a given task variable the cost-minimizing solution is mapped to the origin of the latent space (see Fig.~\ref{fig:title-figure}). In particular:
\begin{itemize}
\item The model is trained on data consisting of configurations and associated task variables and costs, without explicit knowledge of the optimal solution.
\item When trained, approximately optimal inverse kinematic solutions can be generated via a fast operator splitting-based inversion algorithm.
\item Due to the bi-Lipschitz structure of the models, smoothly-varying task variables result in smoothly-varying configuration solutions.
\item We demonstrate the proposed approach finding cost-minimizing configurations for three examples, a simple planar manipulator, a quadrupedal climbing robot and a soft robot manipulator. The results show that it is substantially faster than optimization-based methods, and more accurate than existing generative models.
\end{itemize}

\section{Related Work}

Neural networks have been utilized for inverse problems for many years, with their application to inverse kinematics as early as 1992~\cite{demers_global_1992}, where they used neural networks as approximate mappings between task and configuration space for each kinematic solution group. Subsequently, statistical learning methods such as locally weighted linear regression have been applied~\cite{dsouza2001learning}. More broadly, data-driven methods have been applied to inverse problems across a wide range of scientific and engineering domains, as surveyed in ~\cite{arridge_solving_2019}. 

Generative modeling techniques such as normalizing flows were first applied to inverse kinematics  in~\cite{ardizzone_analyzing_2019} and further developed~\cite{ames_ikflow_2022}, which added noise injection during training to promote greater coverage of the configuration space. Graph-based methods such as \textit{Generative Graphical Inverse Kinematics}~\cite{limoyo_generative_2025} and \textit{GraphDiff-IK}~\cite{huang_whole-body_2026} have demonstrated the effectiveness of networks that simultaneously learn the kinematic topology of the robot and the task space mapping. Whilst these methods collectively produce sets of kinematically feasible solutions, there are no measures included that ensure these solutions are of high quality, e.g. minimizing a user-defined cost function.

Incorporating solution quality for learned kinematics extends the problem to a form of constrained optimization, which can be challenging to learn and often requires a surrogate model to approximate~\cite{amos2023tutorial}.~\cite{amos2017input} presented a means of learning a convex surrogate model, which could be efficiently optimized at the expense of poor model fitting. Improving model fitting loss,~\cite{wang_monotone_2024} used a bi-Lipschitz neural network to construct a Polyak--\L{}ojasiewicz (PL) loss function, whose properties enabled better fitting performance whilst still being amenable to gradient-based optimization. Furthermore,~\cite{liang2023low} demonstrated the capability of invertible neural networks for creating homeomorphic mappings between inequality-constrained solution sets to unit balls, enabling efficient sampling of feasible solutions.      

\section{Preliminaries}

Inverse problems such as IK can be formulated as follows: given a known ``forward'' model $f:\mathcal X \to \mathcal Y$ and a known/desired output $y\in \mathcal Y$, compute an $x\in\mathcal X$ such that $y=f(x)$. 

In many practical cases, e.g. IK for redundant manipulators, there generally exists multiple (even infinitely-many) solutions $x$ satisfying $y=f(x)$. In such cases it makes sense to choose the ``best'' among these, as quantified by some auxiliary cost function or regularizer $J(x)$. This can be formalized as a family of constrained optimization problems parameterized by $y$:
\begin{equation}
    \label{equ:problem-statement}
    \begin{aligned}
        \ninput^\star(\task) :=\arg\min_{\ninput \in \mathcal X}   \; &\aux(\ninput)\\
        \textrm{s.t. } \quad &\fwd(\ninput)=\task.
    \end{aligned}
\end{equation}
While such problems can be tackled through standard constrained optimization tools, some applications have characteristics that motivate alternative approaches based on machine learning:
\begin{enumerate}[label=(\alph*)]
    \item The cost function $J$ or the forward model $f$ may be computationally complex, or black-box simulations without available gradients, making real-time optimization impractical.\label{case:comp}
    \item A model may be completely unavailable, and only experimental observations of $x, y$ and $J$ are available, necessitating a data-driven approach.\label{case:exp}
\end{enumerate}

\subsection{Problem Formulation}
Motivated by these applications, we consider a setting in which the available information consists of a data set:
\begin{equation}\label{eq:data}
    \mathcal D = \{x_i, y_i, J_i\}_{i=1}^N
\end{equation}
where  $y_i=f(x_i)$ and $J_i=J(x_i)$ for each $i$. E.g. this data could be collected through extensive off-line computations for case \ref{case:comp} above, or through experiments for case \ref{case:exp}.

The overall goal of this paper is to introduce a class of models which, when trained on the data set $\mathcal D$, can approximate the solution of Problem~\eqref{equ:problem-statement}, i.e., to learn a model $\hat{x}^\star: \mathcal{Y} \rightarrow \mathcal{X}$ such that $\hat{x}^\star(y)\approx x^\star(y)$ for $y\in \mathcal{Y}$.

\subsection{Inverse Kinematics}

An example of Problem~\eqref{equ:problem-statement} is the inverse kinematics of a robot (described as a kinematic tree), where $\ninput=\cfg\in\mathcal{X}\subset \R^n$ is the joint configuration, $\task\in\mathcal{Y}\subset \R^p$ is the desired effector tasks, and $\fwd: \mathcal X \rightarrow \mathcal Y$ is their associated forward kinematics, i.e., $\fwd(\cfg)=\task$. A robot is considered redundant when $n>p$, in which case a feasible task $y\in \mathcal{Y}$ admits a set of solutions $\smm_y=\{q\in \mathcal{X}\mid f(q)=y\}$. The cost $\aux$ for selecting a preferred configuration $q^\star$ from $\smm_y$ could be distance from a rest or reference pose, joint-limit/singularity avoidance terms, manipulability measures, obstacle-avoidance penalties, or their combinations.

\subsection{BiLipNet and PLNet}
\label{sec:bilipnet-and-plnet}
We first introduce some technical machinery that will be used in our approach. A mapping $\blip:\R^{n}\rightarrow \R^{n}$ is said to be $(\mu,\nu)$ \emph{bi-Lipschitz}, with $\nu\geq \mu>0$, if
\begin{equation}\label{equ:bilip_bounds}
    \mu \norm{x_1-x_2}\leq\norm{\blip(x_1)-\blip(x_2)}\leq\nu \norm{x_1-x_2}
\end{equation}
for all $x_1,x_2\in\R^n$, where $\norm{\cdot}$ denotes the Euclidean norm. Such a mapping is invertible, and its inverse $\blip^{-1}:\R^n\rightarrow\R^n$ is $(1/\nu,1/\mu)$ bi-Lipschitz.

\emph{BiLipNet} is a class of deep neural networks $\blip_\theta:\R^n\rightarrow\R^n$ proposed in \cite{wang_monotone_2024} such that, for any learnable parameter $\theta\in\R^M$, the mapping $\blip_\theta$ is $(\mu,\nu)$ bi-Lipschitz, where $\mu$ and $\nu$ can be either user-specified or learned. Moreover, \cite{wang_monotone_2024} develops an efficient operator splitting algorithm for computing the inverse $x=\blip_\theta^{-1}(y)$.

Building on BiLipNet, we can construct a scalar-output neural network, referred to as a \emph{PLNet} \cite{wang_monotone_2024}, of the form
\begin{equation}\label{equ:plnet}
    \pln(\ninput)=\frac{1}{2}\norm{\blip_\theta(\ninput)}^2+b,
\end{equation}
where $b\in\R$ is a trainable parameter. For any $\theta$ and $b$, this network satisfies the Polyak--\L{}ojasiewicz (PL) condition~\cite{polyak1963gradient,lojasiewicz1963topological}
\begin{equation}\label{equ:pl_condition}
    \frac{1}{2}\norm{\nabla_x\mathcal{P}(x)}^2\geq\mu^2\left(\mathcal{P}(x)-\min_{\tilde{x}\in\R^n}\mathcal{P}(\tilde{x})\right),\; \forall x\in\R^n.
\end{equation}
The PL condition is particularly useful in optimization since it is weaker than convexity, but still implies that gradient methods converge to a global minimum with a linear rate \cite{karimi2016linear}. Consequently, PLNet provides a natural architecture for learning potentially nonconvex surrogate loss functions while the resulting landscape is easy to optimize. Furthermore, the bi-Lipschitz property of $\blip_\theta$ ensures that $\mathcal{P}$ has a unique global minimizer $x^\star$, which can be computed via
\begin{equation}\label{equ:plnet_minimizer}
    x^\star=\blip_\theta^{-1}(0),
\end{equation}
and hence $\min_{x\in\R^n}\mathcal{P}(x)=\mathcal{P}(x^\star)=b$.

For computation the inverse $\blip_\theta^{-1}$, we utilize operator splitting algorithms which allow fast computation without the need for gradient descent \cite{wang_monotone_2024}.

In many applications, it is desirable to preserve these structural properties with respect to the input variable while allowing the network to depend on an additional conditioning variable. To this end, \cite{wang_monotone_2024} also introduces a \emph{partially bi-Lipschitz} neural network $\blip_\theta:\R^n\times\R^p\rightarrow\R^n$ such that, for every model parameter $\theta\in\R^M$ and conditioning variable $p\in\R^p$, the mapping $\blip_\theta(\cdot;p):\R^n\rightarrow\R^n$ is bi-Lipschitz. Accordingly, we can construct a partially PL network as
\begin{equation}\label{equ:conditional_plnet}
    \mathcal{P}(x;p)=\frac{1}{2}\norm{\blip_\theta(x;p)}^2+b_\phi(p),
\end{equation}
where $b_\phi:\R^p\rightarrow\R$ is a scalar-output neural network parameterized by $\phi$. It is easy to obtain that for any conditional variable $p\in \R^p$, the surrogate loss model $\mathcal{P}(\cdot; p)$ has a global minimum $b_\phi(p)$, which is achieved as $x^\star(p)=\blip_\theta^{-1}(0; p)$.

\section{Methodology}

Here we present the main ideas of LOInK, an approach to learn an approximate optimal solution $\hat{x}^\star(y)$ for Problem~\ref{equ:problem-statement} based on the dataset $\mathcal{D} $ in \eqref{eq:data}. We first construct an easy-to-optimize neural surrogate model and then train it based on the supervised dataset $\mathcal{D} $. For online deployment, we take the optimal solution of the surrogate model as an approximation to the true solution of Problem~\eqref{equ:problem-statement}. 

\subsection{Neural Surrogate Model}\label{sec:pseudo-invertible-bilipnet}

Our neural surrogate model takes the form of 
\begin{equation}\label{eq:surrogate-prob}
    \begin{split}
        \hat{\ninput}^\star(\task) :=\arg\min_{\ninput \in \mathcal \R^n }   \; &\hat{\aux}(\ninput)\\
        \textrm{s.t. } \quad &\hat{\fwd}(\ninput)=\task
    \end{split}
\end{equation}
where $y$ is the desired task to achieve. To simplify the presentation, we omit the dependency of $\hat{J}$ and $\hat{f}$ on the learnable  parameter $\theta$. 

Here we give step-by-step construction for $\hat{f}$ and $\hat{J}$. First, we take the following neural coordinate transformation $x\mapsto (\hat{y},z)$ with $\hat{y}\in \R^{p}$ and $z\in \R^{n-p}$ via
\begin{equation}
    \begin{bmatrix}
        \hat{y} \\ z
    \end{bmatrix}=\begin{bmatrix}
        \blip_1(x) \\ \blip_2(x)
    \end{bmatrix}:=\blip(x),
\end{equation}
where $\blip$ is a BiLipNet. Here we choose $\hat{f} = \blip_1$. Secondly, we use the auxiliary variable $z$ to form the cost function
\begin{equation}\label{equ:auxiliary-cost-definition}
    \hat{J}(z, \hat y)=\norm{z}_{M(\hat y)}^2 + b(\hat y)
\end{equation}
where the metric $M(\hat y)\succ 0$ and the offset $b(\hat y)\in \R$ are generated by a neural network
\begin{equation}
    \mathcal{H}: \hat y\mapsto (b, P)
\end{equation}
with $P\in \R^{p\times p}$ and 
$ M=\frac{1}{2}I + P^\top P$.

In summary, the surrogate optimization model in \eqref{eq:surrogate-prob} can be written as 
\begin{equation}\label{eq:surrogate-opt-prob}
    \begin{split}
        \min_{x\in \R^n} \quad  &\norm{\blip_2(x)}_{M(\blip_1(x))}^2+b(\blip_1(x))\\ \text{s.t.} \quad &\blip_1(x)=y.
    \end{split}
\end{equation}
It is obvious that for a given $y$ \eqref{eq:surrogate-opt-prob} admits a unique global minimizer over $x$:
\begin{equation}\label{eq:minimizer}
    x^\star =\blip^{-1}\left(\begin{bmatrix}
        y \\ 0
    \end{bmatrix}\right),
\end{equation}
which can be computed via the fast splitting-based method in \cite{wang_monotone_2024} and used as an approximate solution to Problem~\eqref{equ:problem-statement} after model training.

\subsection{Model Training}
\begin{figure*}
    \centering
    \includegraphics[width=0.8\linewidth]{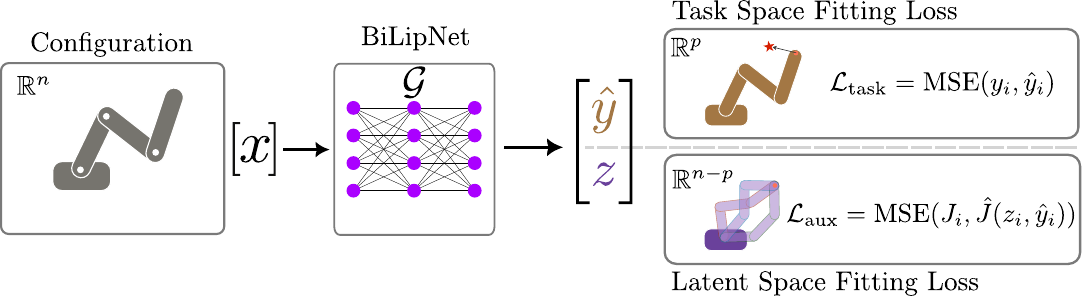}
    \caption{LOInK model structure and training process, where we have highlighted the decoupled output $[\task\; \nll]^T \in \mathbb{R}^n$ and the separate loss functions for each output space. In the task space we impose a fitting loss between dataset tasks $y_i \in \mathbb{R}^p$ and network output $\hat y_i := \hat y(x_i)$ whilst the latent space fitting loss structures the latent space so that the mapping $\hat \aux(z_i, \hat{y}_i)$ defined in~\eqref{equ:auxiliary-cost-definition}  approximates the auxiliary costs $\aux_i$ from the dataset $\data$. 
    }
    \label{fig:training}
\end{figure*}

The training objective is to make the surrogate optimization model match the original problem \eqref{equ:problem-statement} over the dataset $\mathcal{D}$, see Fig.~\ref{fig:training}. Let $\theta$ and $\phi$ denote the model parameters of BiLipNet $\blip$ and neural network $\mathcal{H}$, respectively. 

The first loss component is to minimize the task error:
\begin{equation}\label{eq:task-loss}
    \mathcal{L}_\text{task}(\theta; \data) =\mathrm{MSE}(\hat{y}, y) =\frac{1}{N}\sum^N_{i=1}\norm{\blip_{1}(\ninput_i;\theta)-\task_i}^2_2.
\end{equation} 
The second component, denoted as the \emph{auxiliary} loss, fits our surrogate cost model~\eqref{equ:auxiliary-cost-definition} to the observed costs $J_i$: 
\begin{equation}\label{eq:aux-loss}
    \mathcal{L}_\text{aux}(\theta,\phi;\mathcal D) = \mathrm{MSE}(\hat{J},J)=\frac{1}{N}\sum^N_{i=1}\bigl(\hat{J}(z_i, \hat{y}_i;\theta,\phi)-\aux_i\bigr)^2.
\end{equation}
The total training loss is then given by
\begin{equation}\label{equ:total-loss}
\mathcal{L}(\theta,\phi;\mathcal D) =  \mathcal{L}_\text{task}(\theta;\mathcal D) + \lambda\mathcal{L}_\text{aux}(\theta,\phi;\mathcal D),
\end{equation}
with $\lambda > 0$ as the weighting coefficient.

\subsection{Properties and Geometric Interpretation} 
\label{sec:geometric-interpretation}
The main idea of our approach is to learn a diffeomorphism $\mathcal G$ between configuration space $\mathcal{X}$ and a decoupled coordinate system $\mathcal{Y}\times \mathcal{Z}$, where $\mathcal{Y}$ and $\mathcal{Z}$ denotes the task space and latent space, respectively, as illustrated in Fig.~\ref{fig:title-figure}.

For a given task pose $y\in \mathcal{Y}$, this implies that $\mathcal G$ maps the constraint set $\mathcal{S}_y = \{x\in \mathcal{X}\mid f(x) = y\}$, which may be a curved manifold, onto the ``flat'' space $\{y\} \times \mathcal Z$. This implies $\blip_1(x)=y$ for $x\in \mathcal{S}_y$, motivating the task-space fitting loss. Technically, this decoupling requires that the forward  map $f(x)$ is a submersion defining a foliation over $\mathcal X$ which is trivializable over the desired region. While this may not always be true it is always true in a region of a non-degenerate point \cite[Sec. IV.3]{camacho1985geometric}.

Secondly, the diffeomorphism $\blip$ maps the cost function $J(x)$ over the set $\mathcal{S}_y$ into a positive-definite quadratic form on $ \mathcal Z$ with the global minimum rooted at $z=0$, i.e. $\hat{J}$ in \eqref{equ:auxiliary-cost-definition}. This implies that the optimal solution $\ninput^\star(y) \in \mathcal S_y$ maps to the point $[y, 0]^T \in \{y\} \times \mathbb Z$. It is worth to mention that  $x^\star(y)$ does not need to be included in the training dataset $\mathcal{D}$, instead, it is learned by fitting the auxiliary cost model $\hat{J}$ to the overall loss landscape shape captured by the dataset $\mathcal{D}$.  

Note that we generally do not require the original problem \eqref{equ:problem-statement} possess the same nice property as the neural surrogate model \eqref{eq:surrogate-opt-prob}, which admits a unique minimizer \eqref{eq:minimizer}. Similar to PLNet \cite{wang_monotone_2024}, imposing such a property on the surrogate model can instead be seen as a form of regularization: it encourages the model to disregard spurious poor local minima and focus on regions that are more likely to contain a high-quality solutions. 

\subsection{The Effect of Bi-Lipschitz Bound}\label{sec:velocity_bound}
The bi-Lipschitz bound $(\mu,\nu)$ of the neural network $\blip$  controls the smoothness of the learnt optimal IK solution. Consider the case where the task pose follows a smooth trajectory $y(t)$, then the associated IK solution \eqref{eq:minimizer} also follows a smooth trajectory $x^\star(t)$ satisfying
\begin{equation}
    \frac{1}{\nu}\norm{\dot{y}(t)}\leq \norm{\dot{x}^\star(t)}\leq \frac{1}{\mu}\norm{\dot{y}(t)}\quad \forall t\geq 0.
\end{equation}
In other words, the hyper-parameters $\mu,\nu$ control the velocity bounds in the configuration space, in particular larger $\mu$ leads to smoother configuration trajectories $x^\star(t)$ for a given $y(t)$.

\section{Illustration on a 3-DoF Manipulator} \label{sec:3dof}

We first illustrate our approach on a planar three degree-of-freedom (DoF) manipulator for tracking a horizontal target while minimizing an objective associated with the kinematics of the system (i.e. an instance of case~\ref{case:comp} for Problem~\eqref{equ:problem-statement}). The manipulator has links of uniform length $l = 1$\,m, and each joint restricted to the domain $[0, \pi]$ to  capture only ``elbow up'' configurations.

\subsection{Problem Formulation}
Considering Problem~\ref{equ:problem-statement}, we set the input as the configuration of the robot (i.e. $\ninput = \cfg\in \mathcal C \subseteq \R^3$). Defining the horizontal position of the end-effector relative to the base as $p \in \mathbb R$, the forward model for our problem is the forward kinematics $f(x) = p := y$, mapping from configuration space to this one-dimensional task space. 

The cost function $J$ in Problem~\ref{equ:problem-statement} includes two terms. The first ensures the system remains within its range of motion, via a log-barrier cost on joint limits, normalized to the range $\cfg_i \in [-1, 1]$, the cost is
\begin{equation}
\aux_\cfg(\cfg) = -\sum_i\log(\max(1-\cfg_i^2,\epsilon)),
\label{eq:joint_dev}
\end{equation}
for some small conditioning scalar $0 \le \epsilon \ll 1$.

The second cost relates to the projection of the system's center of mass $c \in \mathbb R^2$ along the horizontal plane, by keeping this projection close to the base of the robot, we treat this cost as a proxy for minimizing the motor torques require to counteract gravitational effects. This cost is computed as
\begin{equation}
\aux_\textrm{com}(\cfg) = \frac{l}{3}\sum_{i=0}^3 \norm{c_{\textrm{com}, x}^i(\cfg)}^2.
\label{eq:com_dev}
\end{equation}
Then, the total cost is
\begin{equation}
    \aux(\cfg) = \aux_\cfg(\cfg) + \lambda_\textrm{com} \aux_\textrm{com}(\cfg),
\end{equation}
where $\lambda_\textrm{com}>0$ is a weighting coefficient.

\subsection{Training Details}

We generate the training dataset $\mathcal D$ by uniformly sampling the configuration space $\mathcal C$ for 1,000,000 samples, and for each configuration sample $x_i = \cfg_i$, compute $\task_i = \fwd(x_i)$ and cost $\aux_i = \aux(x_i)$. The BiLipNet $\mathcal{G}$ consists of two monotone layers \cite{wang_monotone_2024} with depth of 4 and width of 256, trained for 3 epochs with a batch size of 500. The offset $b$ was parameterized by a 2-layer MLP with width of 128. We fix the bi-Lipschitz lower bound $\mu$ at 0.1, keeping the upper bound $\nu$ as a free parameter, but with a small penalty on the distortion $\tau:=\nu/\mu$ in the training loss. Our learning rate was 1e-3, with a reduce on plateau patience of 500 steps.

For this and all following examples, numerical experiments were performed on an Intel Core i9-14900KF CPU with 64GB of RAM and an NVIDIA GeForce RTX 4090 GPU.

\subsection{Results}
\begin{figure}
    \centering
    \includegraphics[width=0.95\linewidth]{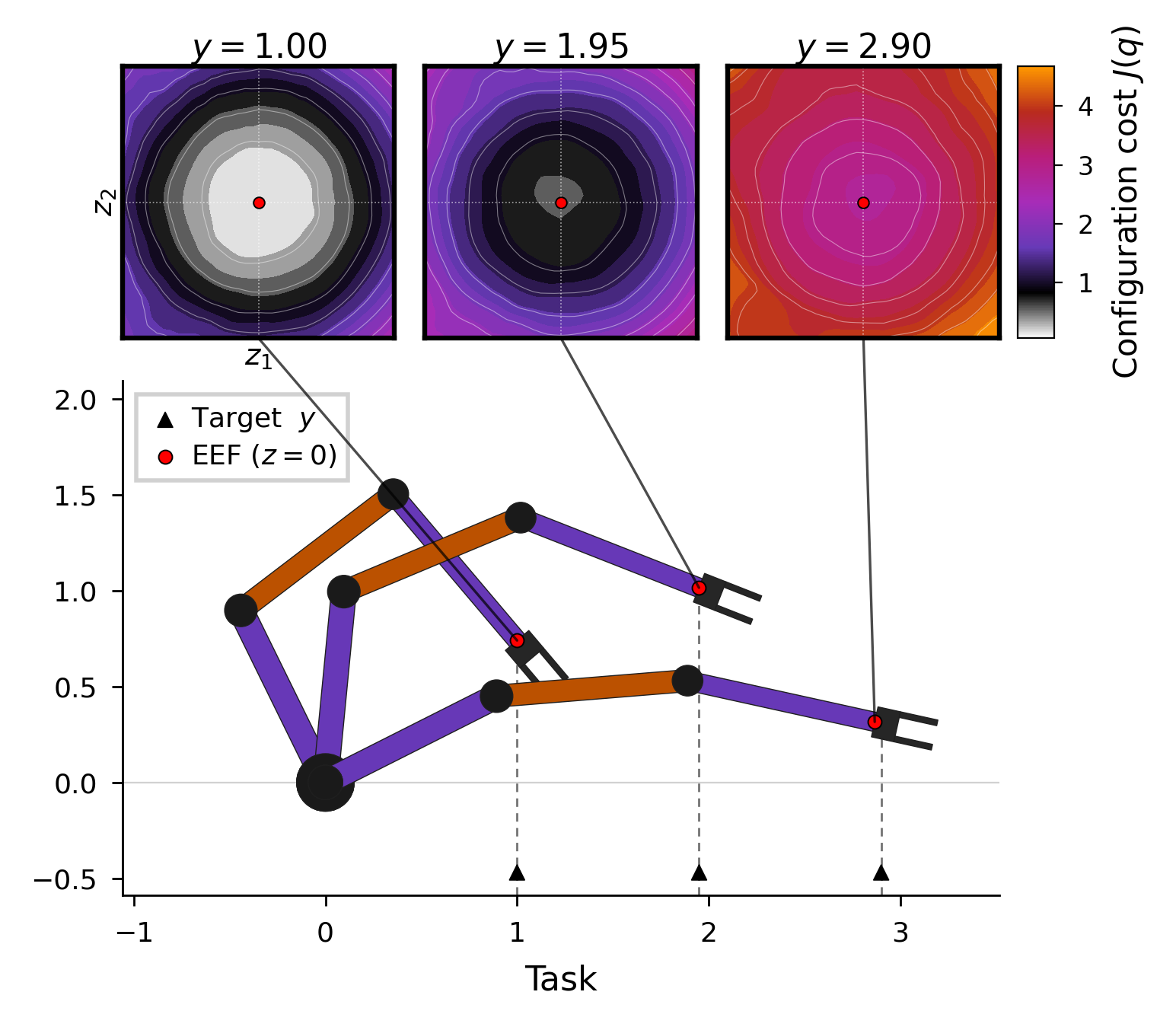}
    \caption{Demonstrative example of the planar 3-DoF manipulator, where the task space is the horizontal position. At each of three task space locations, there is a particular minimal cost. In the learned latent space $z_1, z_2$, cost increases radially from the origin.}
    \label{fig:3dof-manipulator}
\end{figure}
Fig.~\ref{fig:3dof-manipulator} shows sample configurations $\cfg^\star = \blip^{-1}(y)$ for three horizontal targets, with the cost level sets in $\mathcal Z$ shown for each. The influence of $J_\cfg$ is evident with each manipulator remaining well within its joint limits. Furthermore, the case with $y = 1.0\,$m demonstrates the effect of $J_\textrm{com}$, with all links positioned over base of the robot. The properties  of the network discussed in Sec.~\ref{sec:geometric-interpretation} are clearly shown by the circular level sets of $\aux$ at each configuration, with a minimal cost $b(y)$ that varies smoothly with $y$.

\begin{figure}
    \centering
    \includegraphics[width=0.9\linewidth]{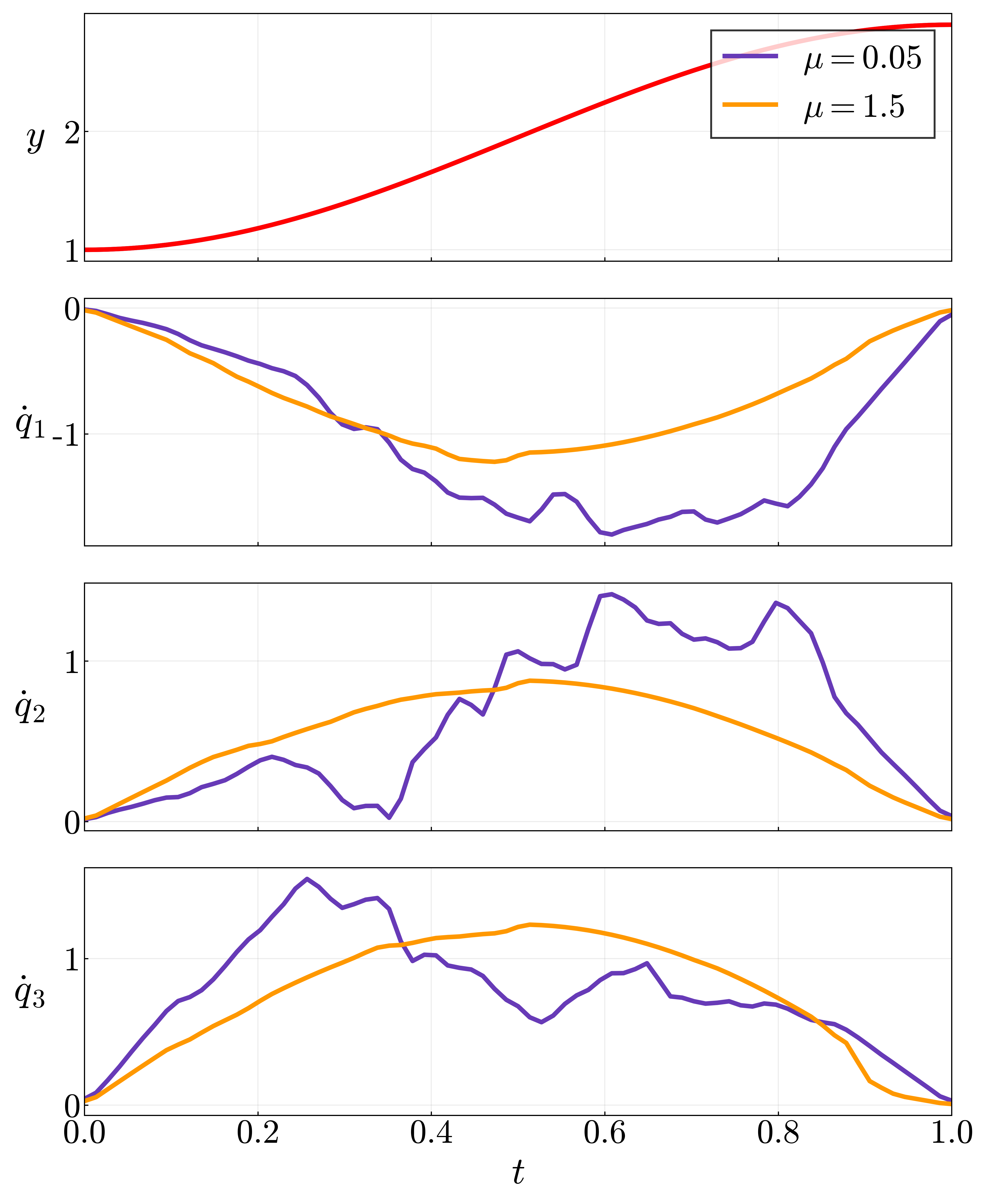}
    \caption{Configuration velocities resulting from end-effector trajectory tracking with different values of $\mu$. We observe that both solutions effectively track the desired task trajectory $y$ (red), while the larger value for $\mu$ produces smoother velocity profiles, as expected c.f. Sec~\ref{sec:velocity_bound}.}
    \label{fig:3dof_vel}
\end{figure}

Fig.~\ref{fig:3dof_vel} illustrates the effects of the bi-Lipschitz bounds $(\mu, \nu)$ for tracking smooth a task trajectory $\task(t)$ (see Sec.~\ref{sec:velocity_bound}). The resulting joint velocities are depicted for two choices of $\mu$. We observe that both solutions effectively track the desired task trajectory $y$, while larger $\mu$ leads to smoother configuration trajectories, as expected.

\section{Application to a Quadrupedal Climbing Robot}\label{sec:magneto-overview}
\begin{figure}
    \centering
    \includegraphics[width=0.95\linewidth]{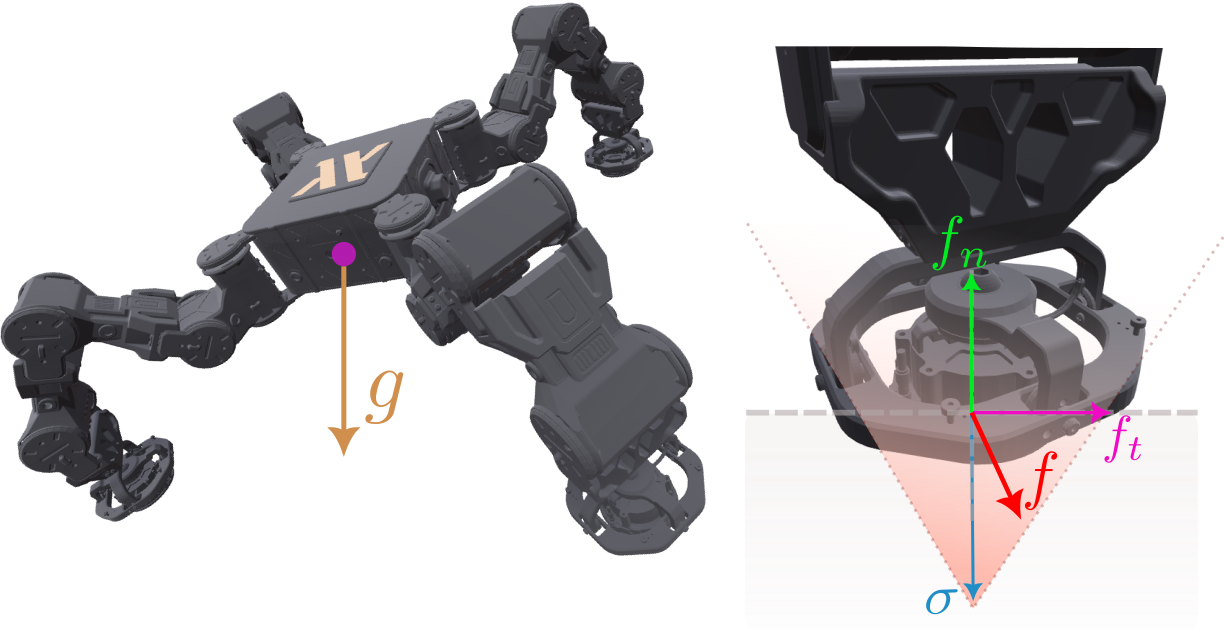}
    \caption{Left: Magneto robot with center of mass and gravity direction. Right: End-effector force $f$ (red) resulting from adhesion force $\sigma$ (blue), frictional normal force $f_n$ (green) and tangential force $f_t$ (magenta).}
    \label{fig:magneto-overview}
\end{figure}

We next demonstrate our approach in finding optimal configurations for a climbing robot, where the underlying objective necessitates the evaluation of a numerical solver, i.e. a high-dimensional example of case~\ref{case:comp} for Problem ~\eqref{equ:problem-statement}.

Magneto~\cite{bandyopadhyay_magneto_2018}, originally developed by CSIRO and now being developed by NEXXIS~\cite{noauthor_designing_nodate}, is a robotic platform designed for inspection tasks in confined and hazardous environments. As shown in Fig.~\ref{fig:magneto-overview}, this robot has $n_l = 4$ limbs, each with four joints and thus $n_j = 16$ joints. Magneto has electromagnetic feet that allow it to climb ferromagnetic surfaces regardless of orientation. It is critical that the forces acting at the contacts are evenly distributed such that no foot exceeds the maximum allowable adhesive force. Under contact-location constraints (the task space), the effector forces can be regulated by shifting the center of mass, by varying the configuration coordinates. 

\subsection{Problem Formulation}
Here we seek to find whole-body configurations $\cfg \in \mathcal C \subseteq \mathbb R^{n_j}$ satisfying contact constraints while distributing the end-effector forces evenly. With reference to Problem~\eqref{equ:problem-statement} and defining $x = \cfg$, we choose as task variable $y$ the set of end-effector positions $p_i(q) \in \mathbb{R}^3$ for $i = 1\hdots n_l$ (each determined via  forward kinematics), i.e.,
\begin{equation}
\task = \begin{bmatrix}
    p_1^\top & p_2^\top & \cdots & p_{n_l}^\top
\end{bmatrix}^\top \in \mathbb{R}^{3 n_l}.
\end{equation}
We design our cost $\aux$ to minimize for the adhesive component of the acting end-effector forces $f \in \mathbb R^{3 n_l}$ required to keep the system in static equilibrium at configuration $\cfg$. As we allow the robot to have any orientation within the inertial frame, we include the perceived gravity vector $g \in \mathbb{R}^3$ (relative to the body frame) as a parameter to the cost. The end-effector forces can be determined by solving a second-order conic program (SOCP) of the form
\begin{equation}
\begin{aligned}
J_f(q, g) = \min_{f, \sigma, s} \quad& s \\
\text{s.t.} \quad& W(q)f = w(q, g)\\
& \norm{f_{i, t}} \leq \mu_f(f_{i, n} +\sigma_i) \quad \forall i\\
& f_{i, n} + \sigma_i \ge 0 \quad \forall i\\
& 0 \leq \sigma_i \leq s \quad \forall i.
\end{aligned}
\label{eq:ad_min}
\end{equation}
With the wrench matrix $W(q) \in \mathbb{R}^{6 \times 3 n_l}$ and bias vector $w(q, g) \in \mathbb{R}^6$ defined in~\cite{bretl_fast_2006} and surface tangent/normal forces given by $f_t$ and $f_n$ respectively. The cost function is given by
\begin{align}
    J(\cfg, g) = J_f(\cfg,g) + \lambda_\cfg \aux_\cfg(\cfg),
\end{align}
with $\lambda_\cfg>0$, where $J_\cfg$ is the joint limit penalty  \eqref{eq:joint_dev}.

Note that in \eqref{eq:ad_min} the normal force constraints allow $f_{i,n}$ to act \textit{into} the surface, up to a maximum adhesion threshold $s \ge 0$ (see Fig.~\ref{fig:magneto-overview}).   In the special case where the maximum adhesion limit $s = 0$,~\eqref{eq:ad_min} is identical to the standard formulation of quasi-static testing with unilateral forces~\cite{bretl_fast_2006}.

\subsection{Training Details}
We train a BiLipNet $\blip$ with the loss~\eqref{equ:total-loss} in order to achieve a mapping from the set of desired contact positions to the adhesion-minimizing configuration $\cfg^\star$. We construct $\blip: \mathcal C \rightarrow \mathcal Y$, with the normalized direction of gravity $\grav \in \mathbb{S}^2$ (relative to the body frame) included as a conditioning variable of $\blip$ (see Sec.~\ref{sec:bilipnet-and-plnet}).

Our network consists of two monotone layers of size 128 with a depth of 2, trained for 10 epochs with a batch size of 500. The offset b was a 2-layer MLP with width of 64. We fix $\mu$ at 0.1, keeping $\nu$ as a free parameter, but with a small distortion penalty in the training loss. Our learning rate was 1e-3, with a reduce on plateau patience of 300 steps.

To generate the training set, we sample $x_i \in \mathcal X,\, \grav_i \in \mathbb{S}^2$ uniformly. For each sample $(\ninput_i, \grav_i)$, we compute the resulting contact task $f(x_i) = y_i$ and cost $J_i = J(x_i; \grav_i)$ by solving Problem~\eqref{eq:ad_min} with CVXPY~\cite{diamond2016cvxpy}.

\begin{figure}
    \centering
    \includegraphics[width=0.95\linewidth]{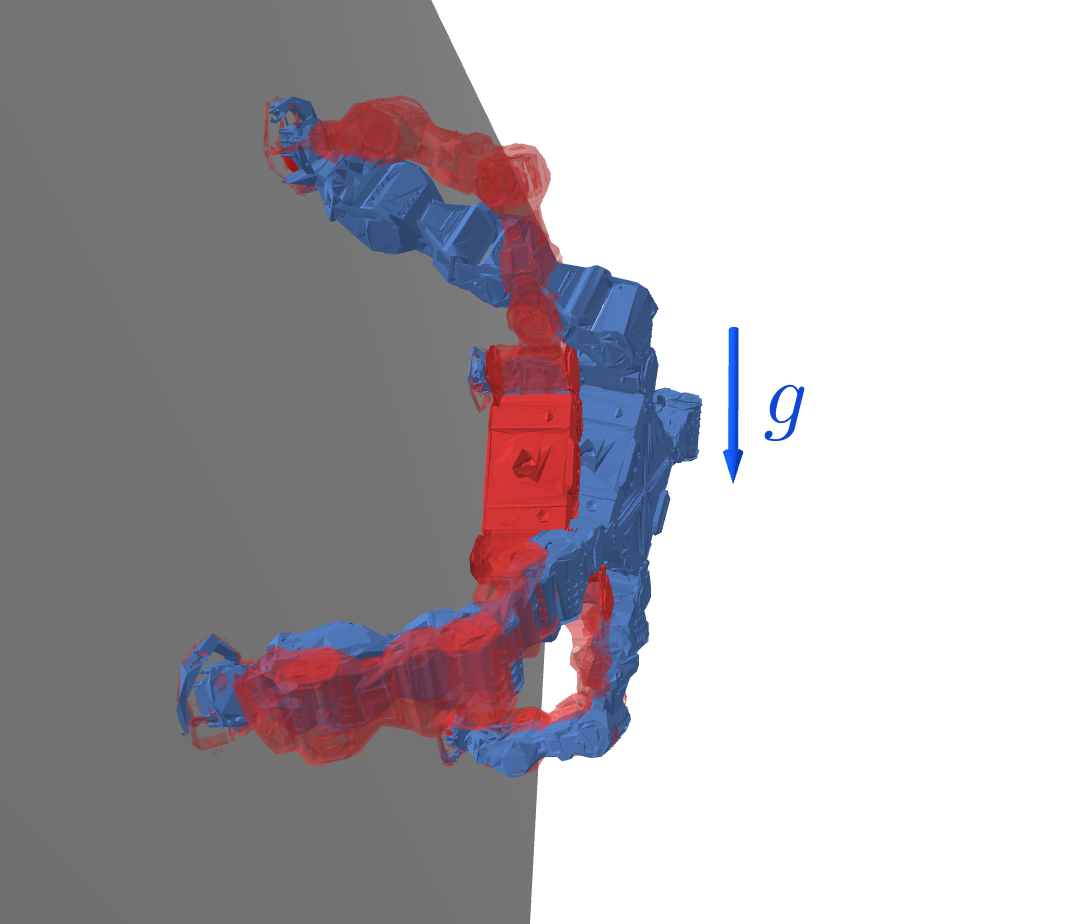}
    \caption{Comparison between a cost minimizing solution (blue) for Magneto  with required adhesion force $J_f = 1.004$\,N and a non-minimizing solution (red) with $J_f = 13.047$\,N.  Note the straight upper limbs of the optimal solution. The direction of gravity is shown with the blue arrow.}
    \label{fig:magneto-ik-solution}
\end{figure}

\begin{table}
\centering
\footnotesize
\setlength{\tabcolsep}{4pt}
\begin{tabular}{lccc}
\toprule
Solver & Time [ms] &  EE error [mm] & Cost ($\aux$)  \\
\midrule
LOInK     & \textbf{5.47} (5.31--5.78)   & 38.9 (17.47--76.5) & 0.237 (0.08--0.46)  \\
IPOPT & 171.1 (77.6--617.7)  & \textbf{15.6} (7.24--30.4)  & \textbf{0.227} (0.08--0.43)\\
\bottomrule
\end{tabular}
\caption{Comparisons between LOInK and optimization method IPOPT [mean (min-max)].}
\label{tab:mag_comparison}
\end{table}

\subsection{Results}
To evaluate the proposed approach we compare against a model-based optimization approach using IPOPT~\cite{wachter2006implementation}. Table \ref{tab:mag_comparison} shows that LOInK generates approximately cost-minimizing solutions on average 31 times faster than the IPOPT solver, and in some cases over 100 times faster. It is also notable that the computation time for LOInK is very consistent, whereas for IPOPT computation time varies by almost an order of magnitude for different task variables. Both methods return solutions of comparable cost, however it is notable that IPOPT achieved lower end-effector error although the errors achieved by LOInK would still be acceptable in most applications.

\section{Application to a Soft Robot Manipulator} 
\label{sec:soft-robot-manipulator}

\begin{figure}
    \centering
    \includegraphics[width=0.9\linewidth]{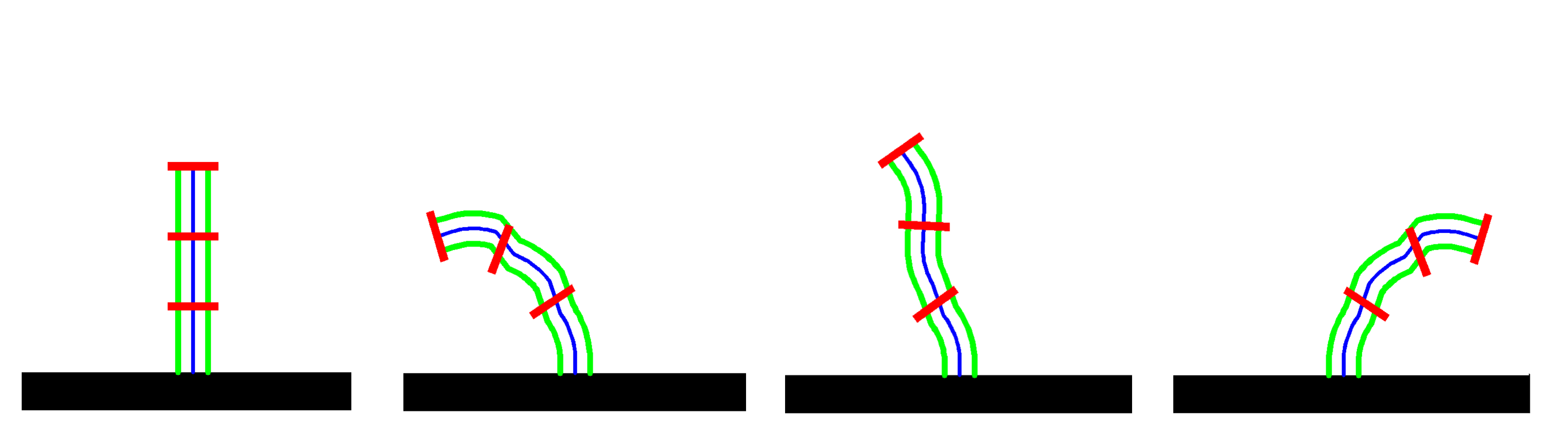}
    \caption{Example postures for the simulated three-segment planar HSA robot, showing configurations that can be achieved on the platform through its actuation ranges.}
    \label{fig:hsa-posture-comparison}
\end{figure}

We now apply the proposed approach to a planar HSA soft manipulator~\cite{paine_design_2014}, where the forward kinematic model is unknown. Instead, we have access to a set of experimental data \cite{fang_efficient_2022}, consisting of the input actuator forces and resulting configurations generated by a complex simulation model, see example postures in Fig.~\ref{fig:hsa-posture-comparison}. This is an example of case~\ref{case:exp} for Problem~\eqref{equ:problem-statement}. 

\subsection{Problem Formulation}
Here we model an $N$-segment manipulator in 2D space within the~\texttt{SoRoMox} framework~\cite{stolzle_soromox_2026}. Each segment contains two actuators which allow segment-wise elongation and lateral bending. We define the control for each segment as $u_{i} := [u_{i, 1}\; u_{i, 2}]^T \in \mathbb{R}^2\; i = 1 \hdots N$, giving us our input $\ninput = u = \begin{bmatrix}
    u_1 & \cdots & u_N
\end{bmatrix}^\top \in \mathcal U \subset \mathbb{R}^{2N}$ and select the position and orientation of the final segment as our task output (i.e. $y = \begin{bmatrix}
    p_x & p_y & \theta
\end{bmatrix}\in \mathcal{Y}\subset \R^{3}$). We treat the forward map $f(x) = y$ in our problem as the evaluation of the complex \texttt{SoRoMox} model. We design our objective to encourage the actuation close to a desired level $\bar{u} \in \mathbb{R}^{2N}$, which we express as a sum of squares cost
\begin{align}\label{eq:J-u}
    \aux(u) = \norm{u - \bar{u}}^2_2.
\end{align}

\subsection{Training Details}

Our BiLipNet $\blip$ consisted of 2 monotone layers, each with 2 hidden layers of width 256. The offset function $b(y)$ had 2 layers of width 64. We chose a learning rate of 1e-3, with a reduce on plateau  patience of 500 steps. For the following experiments we selected a manipulator with $N = 3$, with each segment 0.1\,m in length.

To generate the training dataset $\mathcal D$, we uniformly sampled $2\times 10^6$ inputs $u^i \in \mathcal{U}$ where for each $u^i$, we simulated the corresponding~\texttt{SoRoMox} model, recording the end-effector placement $y_i $ after reaching equilibrium. We also evaluated the cost $J_i = \aux(u^i)$ from \eqref{eq:J-u} with $\bar u = 0$, i.e., the cost becomes effort-minimization. We compare our approach to \textit{IKFlow}~\cite{ames_ikflow_2022}, which was trained on the same dataset with similar model size.

\subsection{Results}

\begin{figure}[!tb]
    \centering
    \includegraphics[width=\linewidth]{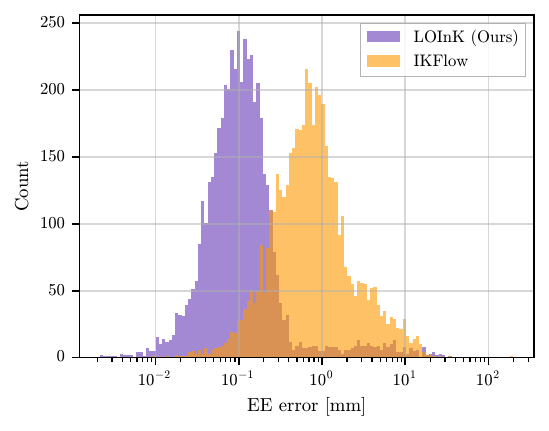}
    \caption{Histogram comparing positional error of the task space position across 2000 end-effector positions.}
    \label{fig:hsa_accuracy_comp}
\end{figure}

\begin{figure}[!tb]
    \centering
    \includegraphics[width=\linewidth]{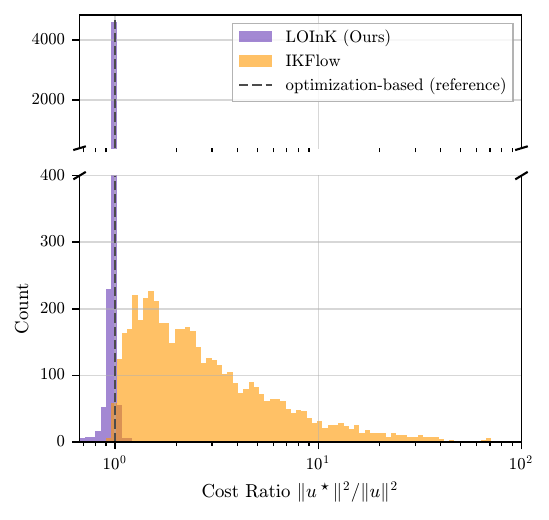}
    \caption{Histogram showing the ratio of actuation efforts versus an optimization-based reference, where our solutions are obtained at $z=0$ for a sample of 2000 random end-effector targets.}
    \label{fig:hsa_cost}
\end{figure}

\begin{figure}[!tb]
    \centering
    \includegraphics[width=\linewidth]{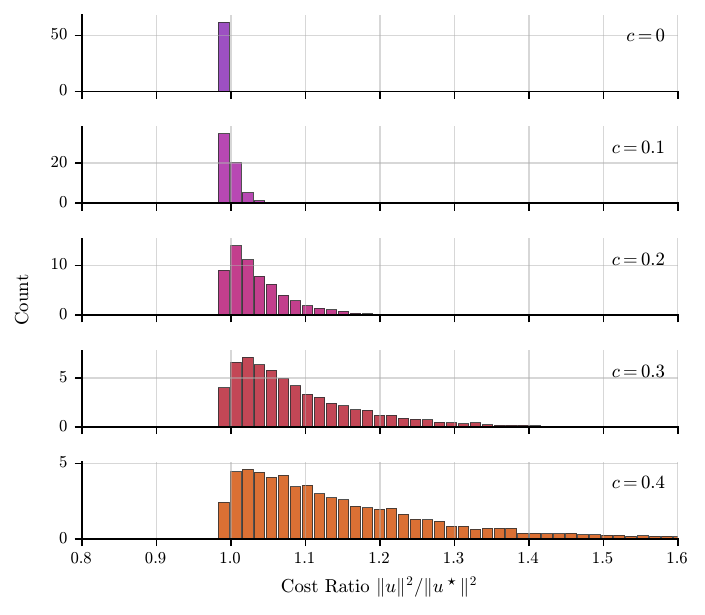}
    \caption{Costs of the distribution obtained by sampling $z\sim \mathcal N(0,cI)$ with varying $c$.}
    \label{fig:hsa_z_sample}
\end{figure}

\begin{table}[!tb]
\centering
\footnotesize
\setlength{\tabcolsep}{4pt}
\begin{tabular}{lccc}
\toprule
Solver & Time [ms] & EE error (mm) & Cost ($J$) \\
\midrule
LOInK & \textbf{2.66} (2.66-2.67) & \textbf{0.44} (0.0022-35.049) & \textbf{0.17} (0.0003-0.59)\\
IKFlow & 4.51 (4.43-4.56) & 1.42 (0.0037-43.80) & 0.38 (0.01-0.71)\\
\bottomrule
\end{tabular}
\caption{Comparison with IKFlow for the Soft Robot Manipulator: Solve times, EE error, and cost [mean (min-max)]}
\label{tab:hsa}
\end{table}

The results in Table~\ref{tab:hsa} show that our method generates solutions faster than IKFlow, with higher end-effector accuracy and lower actuation effort cost. The histogram in Fig.~\ref{fig:hsa_cost} shows that we can directly sample approximately cost-minimizing inputs, as opposed to IKFlow,  which generates diverse samples which must be post-processed to assess quality. This would require generating far more samples and, in a model-free context, would require some kind of learnt model to estimate the resulting costs. In Fig.~\ref{fig:hsa_accuracy_comp} we see that this cost minimization has not come at the cost of accuracy of end-effector position, in fact our method is substantially more accurate. 

If some diversity of solutions is required, we can trade off cost vs diversity by sampling from a distribution centred on $z=0$ with varying covariance, see Fig.~\ref{fig:hsa_z_sample}, where for a particular configuration we  sample $z$ from a zero mean Gaussian with variance $\Sigma=cI$ with varying $c$. 

\section{Limitations}
Our work does have limitations, which could be addressed in future work. Firstly the method assumes that the latent space consists of a single connected component. This is not always the case in inverse kinematics, but could be addressed by learning multiple maps, one per connected component.

Secondly, as with any learning based approach, with limited data and model capacity the precision of our end-effector placement may not be as high as with a model-based method, and as such a solution may need to be refined after being obtained. Finally, the data sampling suffers from the curse of dimensionality, becoming intractable to densely sample the configuration space, and more targeted sampling may be required. 

\section{Conclusions}
In this work we have introduced Learned Optimal Inverse Kinematics ($\methodnameacr$), which learns an invertible mapping from configurations to a decoupled task/latent space, with the latent space structured to enable efficient generation of (approximately) optimal solutions. The bi-Lipschitz structure of the underlying invertible map also guarantees that smoothly-varying task variables lead to smoothly-varying configuration solutions. In application examples, it was demonstrated that the method can generate near-optimal solutions much faster than model-based optimization, and can generate higher-quality and more accurate solutions than existing learning-based approaches.

\section{Acknowledgments}
This research was supported by the Australian Research Council through the ARC Research Hub in Intelligent Robotic Systems for Real-Time Asset Management (IH210100030) and by NEXXIS PTY LTD. We would also like to thank NEXXIS for access to the Magneto simulation model and useful discussions regarding Magneto's operation.

\bibliographystyle{IEEEtran}
\bibliography{bibliography}

\end{document}